\documentclass[acmsmall, manuscript]{acmart}

\makeatletter
\@ifundefined{Bbbk}{}
  {}
\makeatother

\usepackage{amsmath,amssymb}
\usepackage{algorithm}
\usepackage[noend]{algpseudocode}
\usepackage{float}
\usepackage{graphicx}
\usepackage{tabularx, booktabs, multirow}
\usepackage{bm}
\usepackage{caption}
\usepackage{tikz}
\usetikzlibrary{arrows.meta, positioning, fit}

\newcolumntype{Z}{>{\centering\let\newline\\\arraybackslash\hspace{0pt}}X}

\title[Attention-Based Spatial-Temporal Fusion GCN]{Eliminating Propagation Delay: Attention-Based Spatial-Temporal Fusion Graph Convolution Network for Traffic Flow Prediction}
\author{Jinpeng Chen}
\authornote{Corresponding author}
\authornote{Also with the Key Laboratory of Trustworthy Distributed Computing and Service (Beijing University of Posts and Telecommunications), Ministry of Education, Beijing, China.}

\affiliation{%
  \institution{School of Computer Science (National Pilot Software Engineering School), Beijing University of Posts and Telecommunications}
  \city{Beijing}
  \country{China}}
\email{jpchen@bupt.edu.cn}

\author{Ziyu Yu}
\authornotemark[2]
\email{2022140861@bupt.edu.cn}
\affiliation{%
  \institution{School of Computer Science (National Pilot Software Engineering School), Beijing University of Posts and Telecommunications}
  \city{Beijing}
  \country{China}
}

\author{Tao Wang}
\authornotemark[2]
\email{kkkwt1998@163.com} 
\affiliation{%
\institution{School of Computer Science (National Pilot Software Engineering School), Beijing University of Posts and Telecommunications}
  \city{Beijing}
  \country{China}
}

\author{Jun Ma}
\authornotemark[2]
\email{majun@bupt.edu.cn}
\affiliation{%
  \institution{School of Computer Science (National Pilot Software Engineering School), Beijing University of Posts and Telecommunications}
  \city{Beijing}
  \country{China}
}

\author{Hongbo Gao}
\email{ghb48@ustc.edu.cn}
\affiliation{%
  \institution{Department of Automation, School of Information Science and Technology, University of Science and Technology of China}
  \city{Hefei}
  \country{China}
}

\author{Senzhang Wang}
\email{szwang@csu.edu.cn}
\affiliation{%
  \institution{School of Computer Science and Engineering, Central South University}
  \city{Changsha}
  \country{China}
}

\author{Zufeng Zhang}
\email{zhangzufeng@tsari.tsinghua.edu.cn}
\affiliation{%
  \institution{Department of Automation, Tsinghua University}
  \city{Beijing}
  \country{China}
}

\author{Kaimin Wei}
\email{kaiminwei@jnu.edu.cn}
\affiliation{%
  \institution{College of Information Science and Technology, Jinan University}
  \city{Guangzhou}
  \country{China}
}

\begin{document}

\begin{abstract}
Predicting traffic flow is crucial to optimizing transportation systems and improving urban mobility. Many graph convolution-based models have been proposed to extract spatial-temporal features and predict traffic flow. However, most focus on spatial-temporal and semantic correlation in topological relationships. There are two primary problems to address. Firstly, the convolutional structure in the model focuses on utilizing static spatial dependencies and spatial-temporal relationships in topological structures, while neglecting the different information propagation delays between adjacent nodes in the convolution. Secondly, these methods often stack a large number of complex structures, resulting in a substantial increase in computational time during the model training phase, thereby disregarding the model's requirements for timeliness. In this paper, we propose a novel network called the Attention-Based Spatial-Temporal Fusion Graph Convolution Network (A-STFGCN). We design a spatial-temporal fusion block to extract the spatial-temporal feature correlations with propagation delay errors removed and to capture both long-term and short-term temporal characteristics of the data within a multi-head self-attention mechanism based on a mask matrix. Extensive experiments on five real-world datasets demonstrate that our method achieves the best overall performance while having good computation and data utilization efficiency compared with the eight baseline methods.
\end{abstract}



\maketitle  

\section{Introduction}
The widespread application of Intelligent Transportation Systems (ITS) in urban traffic management signifies a profound integration of technology into the transportation sector. Combining advanced information technology, communication systems, and data processing techniques, ITS aims to optimize the operation of urban transportation systems, improving efficiency, safety, and sustainability\cite{yin2015literature}. However, the complexity and unpredictability of urban traffic systems still pose challenges, making accurate traffic management a continuous endeavor. Within this dynamic context, the exploration of traffic flow prediction has emerged as an imperative and progressively vital area of study.

\par
Traffic flow, a key indicator of road conditions, is crucial in guiding the management and control of urban traffic. Within the domain of Intelligent Transportation Systems (ITS), precise and reliable traffic flow prediction serves as a valuable tool for route planning and congestion prevention. So far, many deep learning-based short-term traffic flow prediction models have been proposed \cite{xu2022spatial,2024-40113,20241751}. However, most models focus on the topological structure of the entire transportation network in large cities or regions, which means that modeling and training may concentrate more on capturing the temporal pattern of adjacent nodes in the topological structure, ignoring the spatial-temporal heterogeneity in transportation data, making it difficult to accurately capture and model complex and dynamic spatial-temporal correlations in transportation data \cite{yin2021deep}.
\par
For example, in the long term, traffic flow exhibits notable variations at different times of the day. During peak commute hours, road traffic typically experiences a rapid surge, while it tends to be sparser during off-peak hours. Urban nodes with diverse functions often manifest analogous patterns in traffic flow changes, such as congestion in shopping centers during weekends and evenings. On the other hand, traffic conditions are significantly influenced by events. Special events such as sports games, concerts, or emergency road maintenance can have a considerable impact on traffic flow in specific areas and during particular periods. Moreover, due to the varying physical distances between different nodes, the time it takes for traffic information to propagate between nodes often also varies. Existing models and methodologies face challenges in effectively capturing the characteristics of highly dynamic spatial-temporal data.
\par
In the initial phase of traffic data analysis, Convolutional Neural Networks (CNNs) were applied to grid-based traffic data, effectively capturing spatial correlations \cite{zhang2019short,shao2021traffic}. Parallel to this, the temporal dynamics of the traffic were deciphered using Recurrent Neural Networks (RNNs) and Long Short-Term Memory (LSTM) networks \cite{tian2015predicting,du2017traffic,kang2017short,lu2021combined}. This bifurcated approach evolved with the introduction of Graph Neural Networks (GNNs). GNNs, which aligned more closely with the inherent graph structure of traffic networks, offered a more nuanced understanding of spatial dependencies \cite{yu2017spatio}. GNN-based models predominantly model spatial dependencies in a statical way, either predefined \cite{li2021spatial,guo2022fast,song2020spatial} or through self-learning \cite{wu2019graph}. This approach exhibits limitations in comprehending the dynamic and often unpredictable patterns of urban traffic. The static nature of these models fails to capture the temporal variability and evolution inherent in traffic flows. In recent research, there has been a shift towards integrating self-attention mechanisms with GNN models to extract temporal features more accurately \cite{jiang2023pdformer,yan2021learning,ye2022meta,xu2020spatial,velivckovic2017graph}. These models employ different spatial graphs to establish more precise spatial relationships, enhancing the model's ability to predict traffic patterns. However, these GNN-based models with instant messaging mechanisms often neglect the potential impact of time delay in the propagation of spatial information \cite{chen2024pamt}. For example, the time that takes for a traffic incident in one location to affect adjacent areas is frequently overlooked. This aspect underlines the necessity for models that can adequately account for time delays in spatial information propagation, ensuring a more realistic representation of traffic dynamics.

\par
To address the above issues, we propose an Attention-Based Spatial-Temporal Fusion Graph Convolution Network (A-STFGCN) for traffic flow prediction. To facilitate the simultaneous modeling of time dependencies at different granularities, we have designed a novel multi-head temporal self-attention module to capture dynamic spatial correlations. This module integrates local neighborhood information and global information influenced by various factors into self-attention interactions through different graph masking methods, allowing it to capture both short-term and long-term temporal features in traffic data simultaneously. Building on this module, we have further designed a delayed perception feature-based spatial-temporal fusion graph convolution module, calculating temporal correlation matrices with different time steps and revealing hidden spatial-temporal dependencies through fusion graph convolutions of sequential time step data.

\par
In summary, the main contributions of this paper are as follows:
\begin{itemize}
\item{We propose a novel A-STFGCN framework based on the multi-head self-attention mechanism and Spatial-Temporal Fusion convolution for accurate traffic flow prediction. Our approach fully addresses the issues of obtaining Spatial-Temporal relationships in topological structures of traffic data.}
\item{We design a multi-head temporal self-attention module to effectively and efficiently extract and fuse long- and short-range temporal features of the data via different graph masking methods. Then, we propose a spatial-temporal fusion graph module to eliminate information propagation delay errors and synchronously capture spatial-temporal correlations in topological relationships.}
\item{We evaluate the proposed A-STFGCN framework on five real-world traffic speed prediction datasets, which achieves the best overall performance but also has good computational efficiency and data utilization efficiency compared to the eight baseline methods.}
\end{itemize}

\section{Related Work}

\subsection{Spatial-Temporal Traffic Forecasting and Dynamic Graph Learning}
Traffic flow prediction is a classic spatial-temporal forecasting task for intelligent transportation systems. Early deep models usually handled temporal and spatial dependencies with separate components, such as recurrent networks for time series and graph convolution for road topology. Representative GNN-based models, including DCRNN, STGCN, T-GCN, and Graph WaveNet, improved traffic forecasting by propagating information on physical or learned road graphs \cite{li2017diffusion,yu2017spatio,zhao2019t,wu2019graph}. Later, localized and fusion spatial-temporal graph models, such as STSGCN and STFGNN, further improved cross-time graph propagation \cite{song2020spatial,li2021spatial}. However, a fixed road graph or a purely synchronous spatial-temporal graph may still be insufficient when traffic states propagate with different time lags across road segments.

Recent studies therefore increasingly focus on adaptive and dynamic graph learning. DSTAGNN learns dynamic spatial-temporal aware graphs, MVDGCN models multi-view dynamic station relations, and FEDDGCN further decouples traffic signals while enhancing dynamic graph learning with frequency-domain information \cite{lan2022dstagnn,huang2023multi,zhang2025feddgcn}. MetaDG extends dynamic graph modeling by generating dynamic node representations, adjacency matrices, and meta-parameters for spatio-temporal prediction \cite{zou2026meta}. These methods demonstrate the importance of dynamic spatial dependencies, but their learned graph structures are often implicit. In contrast, A-STFGCN explicitly constructs delay-temporal graphs with DTW-based shifted alignment, making the cross-node propagation delay more interpretable and reusable as an offline graph prior.

\subsection{Attention, Transformer and Decomposition-based Forecasting}
Self-attention and Transformer-style models have also become important tools for traffic forecasting because they can capture long-range temporal and spatial dependencies \cite{vaswani2017attention,xu2020spatial,yan2021learning,ye2022meta,reza2022multi}. ASTGCN introduces spatial-temporal attention for traffic forecasting, SAST-GNN applies self-attention to spatio-temporal graph prediction, GDFormer learns graph diffusion attention, and Bi-STAT uses spatial-adaptive and temporal-adaptive Transformer modules \cite{guo2019attention,xie2020sast,su2022gdformer,chen2022bidirectional}. PDFormer is particularly relevant because it studies propagation-delay-aware long-range Transformer modeling \cite{jiang2023pdformer}. Compared with these Transformer-oriented methods, our model uses a lighter GCN-based backbone with masked temporal self-attention: the delay relation is first encoded into explicit fusion graphs, and the attention branch is then used to complement local convolution with DTW-guided temporal dependencies.

Another recent trend is to decompose traffic signals into more interpretable components. STDN disentangles trend-cyclical and seasonal patterns with spatio-temporal embeddings, FEDDGCN separates prominent periodic and perturbative components with frequency filters, and DualCast explicitly separates intrinsic traffic patterns from aperiodic environmental events \cite{cao2025spatiotemporal,zhang2025feddgcn,ijcai2025p366}. These works show that traffic forecasting benefits from modeling heterogeneous temporal patterns rather than treating all observations uniformly. A-STFGCN is complementary to this line of research: instead of decomposing the input series into frequency or event components, it focuses on the propagation pathway by aligning node pairs under short time shifts and injecting the resulting delay-aware relations into graph convolution and temporal attention.

\subsection{Hierarchical and Large-scale Graph Modeling}
Large road networks bring additional challenges because full node-level message passing can be computationally expensive and may obscure regional traffic patterns. Hierarchical models introduce regional or macro-micro graph structures for traffic forecasting, and STAGCN-EC combines attention with graph convolution for efficient traffic forecasting \cite{wang2022hierarchical,guo2021hierarchical,lai2022spatial}. More recently, latent graph structure learning has been proposed for large-scale traffic forecasting by learning adaptive patch assignments in a data-driven manner \cite{wang2025latent}. These approaches motivate the use of coarser spatial abstractions for large-scale scenarios. A-STFGCN follows this direction but keeps the macro graph construction simple and interpretable: spectral clustering produces a static cluster graph, and the transmit block dynamically sends macro-level regional trends back to node-level features.

Table \ref{table:related models} summarizes the above model families from the perspective of the components most relevant to this work. Rather than claiming novelty from self-attention or graph convolution alone, A-STFGCN is positioned as a combined design that integrates macro-micro spatial abstraction, explicit delay-aware graph construction, and DTW-guided masked temporal self-attention.

\begin{table*}
\caption{Comparison of representative traffic forecasting models and recent related studies.}
\centering
\renewcommand{\arraystretch}{1.15}
\setlength{\tabcolsep}{2.5pt}
\scriptsize
\begin{tabular}{l c c c c c} 
\toprule
\multirow{2}{*}{Models} & \multicolumn{5}{c}{Key modeling components} \\ \cmidrule(lr){2-6}
 & GCN & \begin{tabular}[c]{@{}c@{}}Adaptive/\\Dynamic\\Graph\end{tabular} & \begin{tabular}[c]{@{}c@{}}Macro/\\Patch\\Graph\end{tabular} & Delay-aware & \begin{tabular}[c]{@{}c@{}}Attention/\\Decomposition\end{tabular} \\ 
\midrule
ARIMA / GRU & - & - & - & - & - \\
DCRNN / STGCN / T-GCN & $\checkmark$ & - & - & - & - \\
Graph WaveNet / DSTAGNN & $\checkmark$ & $\checkmark$ & - & - & - \\
STFGNN & $\checkmark$ & - & - & $\checkmark$ & - \\
ASTGCN / GDFormer / Bi-STAT & $\checkmark$ & $\checkmark$ & - & - & $\checkmark$ \\
PDFormer & $\checkmark$ & $\checkmark$ & - & $\checkmark$ & $\checkmark$ \\
HGCN / STAGCN-EC & $\checkmark$ & - & $\checkmark$ & - & $\checkmark$ \\
STDN / FEDDGCN / DualCast & $\checkmark$ & $\checkmark$ & - & - & $\checkmark$ \\
Latent Graph Learning & - & $\checkmark$ & $\checkmark$ & - & - \\
MetaDG & $\checkmark$ & $\checkmark$ & - & - & - \\
\midrule
A-STFGCN (ours) & $\checkmark$ & $\checkmark$ & $\checkmark$ & $\checkmark$ & $\checkmark$ \\
\bottomrule
\end{tabular}
\label{table:related models}
\end{table*}

\section{Preliminaries}

\subsection{Notations and Definitions}
\noindent {\bf{Definition 1 (Traffic Road Network):}} We represent the traffic road network as an undirected graph \(\mathcal{G}=(\mathcal{V},\mathcal{E},\bm{A})\), where \(\mathcal{V}=\{v_{1},...,v_{N}\}\) is the set of \(N\) sensor nodes and \(\mathcal{E} \subseteq \mathcal{V}\times\mathcal{V}\) is the set of physical road connections. \(\bm{A}\in \mathbb{R}^{N \times N}\) is the adjacency matrix of \(\mathcal{G}\). In this paper, \(\bm{A}_{ij}=1\) indicates that nodes \(v_i\) and \(v_j\) are directly connected.

\noindent {\bf{Definition 2 (Traffic Graph Signal):}} We use \(\bm{X}_{t}\in \mathbb{R}^{N \times D}\) to denote the traffic observations of all nodes at time step \(t\), where \(D\) is the feature dimension collected by sensors. The historical traffic tensor is denoted by \(\bm{X}=(\bm{X}_{1},...,\bm{X}_{T}) \in \mathbb{R}^{N \times T \times D}\).

\noindent {\bf{Definition 3 (Propagation Delay):}} We estimate the temporal similarity between node pairs using the Dynamic Time Warping (DTW) algorithm. For two nodes \(v_i\) and \(v_j\), if the DTW similarity between the time series of \(v_i\) and the \(k\)-step shifted time series of \(v_j\) is higher than their synchronous similarity, we regard this pair as exhibiting a \(k\)-step propagation delay.

\subsection{Problem Formalization}
Traffic flow prediction aims to predict future traffic states from historical observations on a road network. Given the graph \(\mathcal{G}\), an input window of length \(T_1\), and a prediction horizon of length \(T_2\), our goal is to learn a mapping function \(f\) such that

\begin{equation}
\label{deqn_ex1}
\{\bm{X}_{(t-T_1+1)},...,\bm{X}_{t};\mathcal{G}\}\xrightarrow{f}\{\bm{X}_{(t+1)},...,\bm{X}_{t+T_2};\mathcal{G}\}
\end{equation}

\begin{table*}[!t]
\centering
\caption{Core notations used in A-STFGCN.}
\label{table:notations}
\small
\begin{tabularx}{\textwidth}{l l X}
\toprule
\textbf{Symbol} & \textbf{Shape/Type} & \textbf{Meaning} \\
\midrule
\(\mathcal{G}=(\mathcal{V},\mathcal{E},\bm{A})\) & Graph & Traffic road network with node set, edge set, and adjacency matrix. \\
\(N, N^{C}\) & Scalars & Number of sensor nodes and number of macro nodes in the cluster graph. \\
\(T_1, T_2, D\) & Scalars & Input sequence length, prediction horizon, and feature dimension. \\
\(\bm{X}, \bm{X}^{C}\) & \(\mathbb{R}^{N \times T_1 \times D}\), \(\mathbb{R}^{N^{C} \times T_1 \times D}\) & Node-level input tensor and macro-graph input tensor. \\
\(\bm{Mat}_{T}, \bm{Mat}_{T}^{d}\) & \(\mathbb{R}^{N \times N^{C}}\) & Static cluster assignment matrix and dynamic transmit matrix. \\
\(\bm{\mathcal{X}}\) & \(\mathbb{R}^{N \times T_1 \times 2D}\) & Node representation after fusing node-level and macro-level features. \\
\(\bm{A}_{sg}\) & \(\mathbb{R}^{N \times N}\) & Spatial adjacency matrix derived from the physical road network. \\
\(\bm{A}_{tg}^{1}, \bm{A}_{tg}^{2}\) & \(\mathbb{R}^{N \times N}\) & Delay-temporal graphs corresponding to 1-step and 2-step propagation delays. \\
\(\bm{A}_{tc}\) & \(\mathbb{R}^{N \times N}\) & Temporal self-correlation matrix in the spatial-temporal fusion graph. \\
\(\bm{A}_{stfg}\) & Graph / matrix set & Spatial-temporal fusion graph used by GSTF-GCN. \\
\(\bm{M}_{t}\) & \(\mathbb{R}^{T_1 \times T_1}\) & DTW-based temporal similarity mask for masked temporal self-attention. \\
\(\hat{\bm{Y}}\) & \(\mathbb{R}^{N \times T_2 \times D}\) & Predicted future traffic flow. \\
\bottomrule
\end{tabularx}
\end{table*}

\section{Methodology}
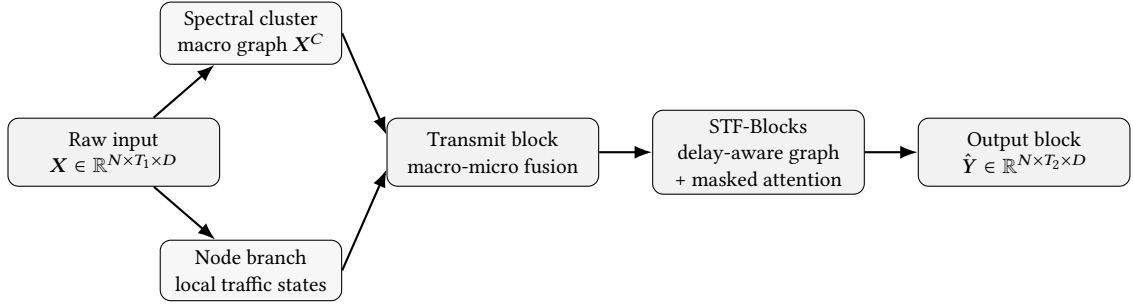
\begin{figure}[t]
\centering
\begin{tikzpicture}[
    node distance=7mm and 7mm,
    >=Latex,
    box/.style={draw, rounded corners, align=center, font=\small, minimum height=9mm, minimum width=28mm, fill=gray!10},
    smallbox/.style={draw, rounded corners, align=center, font=\small, minimum height=8mm, minimum width=24mm, fill=gray!6}
]
\node[box] (input) {Raw input\\$\bm{X}\in\mathbb{R}^{N\times T_1\times D}$};
\node[smallbox, above right=of input, xshift=-15mm] (cluster) {Spectral cluster\\macro graph \(\bm{X}^{C}\)};
\node[smallbox, below right=of input, xshift=-15mm] (nodebranch) {Node branch\\local traffic states};
\node[box, right=22mm of input] (transmit) {Transmit block\\macro-micro fusion};
\node[box, right=of transmit] (stf) {STF-Blocks\\delay-aware graph\\+ masked attention};
\node[box, right=of stf] (output) {Output block\\$\hat{\bm{Y}}\in\mathbb{R}^{N\times T_2\times D}$};

\draw[->, thick] (input) -- (cluster);
\draw[->, thick] (input) -- (nodebranch);
\draw[->, thick] (cluster.east) -- ([yshift=2.3mm]transmit.west);
\draw[->, thick] (nodebranch.east) -- ([yshift=-2.3mm]transmit.west);
\draw[->, thick] (transmit) -- (stf);
\draw[->, thick] (stf) -- (output);
\end{tikzpicture}
\caption{High-level data flow of A-STFGCN. Historical observations are processed by a node-level branch and a macro-graph branch, then fused before the STF-Blocks generate future traffic predictions.}
\Description{A flow diagram showing raw traffic input branching into a spectral clustering branch and a node branch, followed by a transmit block, STF-Blocks, and an output block that predicts future traffic flow.}
\label{fig:pipeline_overview}
\end{figure}

\begin{figure}[t]
\centering
\includegraphics[width=0.95\linewidth]{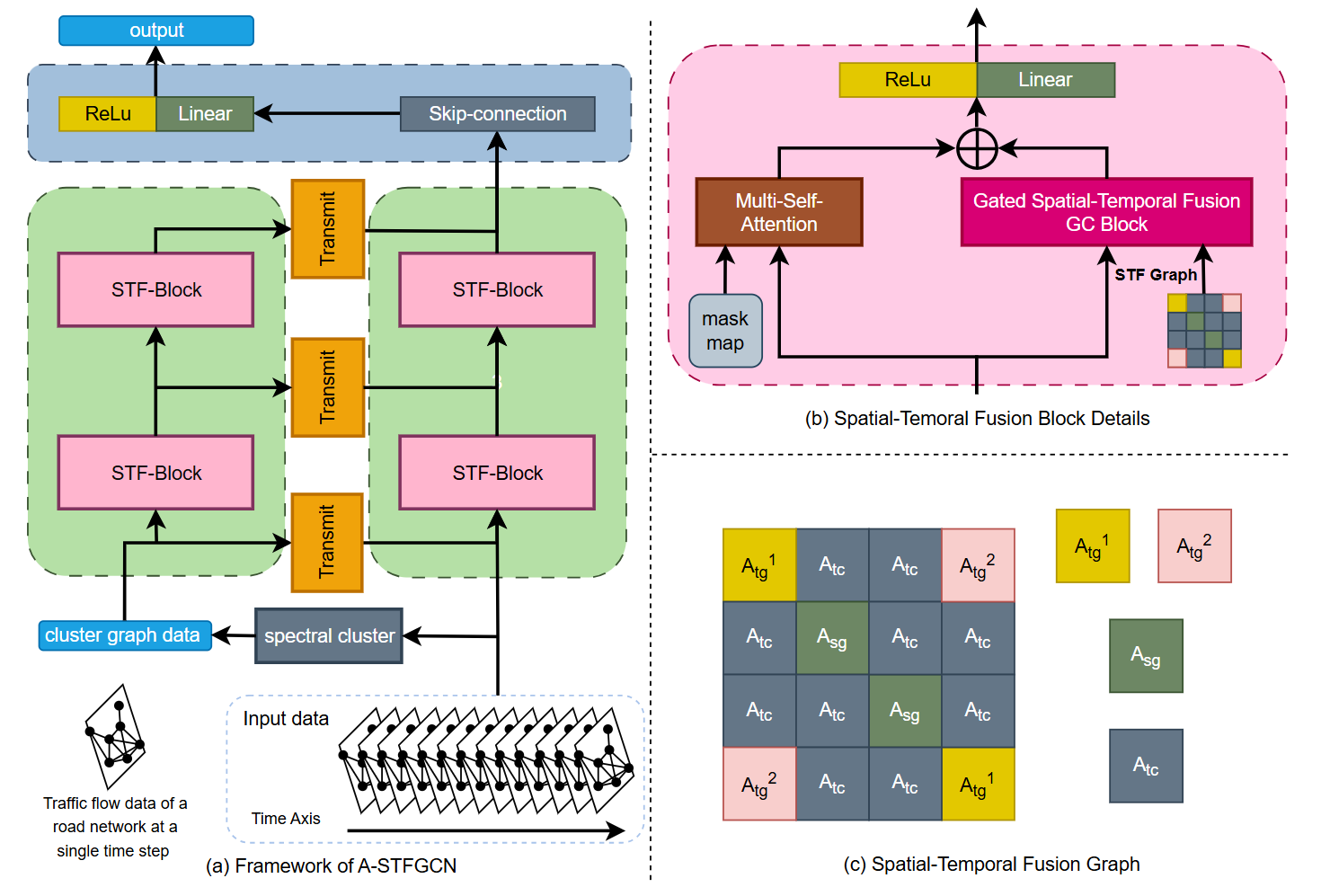}
\caption{Detailed framework of A-STFGCN. (a) shows the parallel node-level and cluster-graph branches. (b) shows the detailed architecture of an STF-Block. (c) illustrates a spatial-temporal fusion graph composed of delay-temporal graphs \(\bm{A}_{tg}^{1}\) and \(\bm{A}_{tg}^{2}\), the spatial graph \(\bm{A}_{sg}\), and the temporal self-correlation graph \(\bm{A}_{tc}\).}
\Description{A detailed architecture diagram of A-STFGCN showing node-level and cluster-level branches, transmit blocks, STF-Blocks, and the spatial-temporal fusion graph components.}
\label{fig:A-STFGCN}
\end{figure}

Figure \ref{fig:pipeline_overview} summarizes the high-level data flow of A-STFGCN, while Figure \ref{fig:A-STFGCN} provides the detailed module-level architecture. Given raw traffic observations \(\bm{X} \in \mathbb{R}^{N \times T_1 \times D}\), the model first constructs a macro graph by spectral clustering, then transfers macro-level features back to the node-level branch through a transmit block, next extracts delay-aware spatial-temporal features with STF-Blocks, and finally outputs \(\hat{\bm{Y}} \in \mathbb{R}^{N \times T_2 \times D}\). Therefore, the core of A-STFGCN is not a simple stack of modules, but the combination of macro-micro spatial abstraction, a delay-aware fusion graph, and masked temporal self-attention.

The original graph branch and the cluster graph branch are processed in parallel. The macro graph provides stable regional context for large-scale road networks, while the node-level branch preserves fine-grained local dynamics. Their interaction allows the model to jointly capture regional propagation patterns and road-segment-level temporal variations.

\subsection{\bfseries Spectral Cluster}
Given the raw input \(\bm{X}=(\bm{X}_{1},\bm{X}_{2},...,\bm{X}_{T_{1}}) \in \mathbb{R}^{N \times T_{1} \times D}\), we first build a macro graph to reduce the complexity of large road networks and to capture propagation patterns at the region-level. This step provides a more stable spatial context for subsequent node-level prediction.

Our motivation is that traffic propagation is usually more coherent within locally connected regions than across the entire city-scale graph. Therefore, adjacent nodes with similar structural roles can be grouped into macro nodes without losing the dominant regional dynamics. This macro representation complements the original graph by emphasizing coarse-grained spatial organization.

\begin{figure}[H]
\centering
\includegraphics[width=0.6\linewidth]{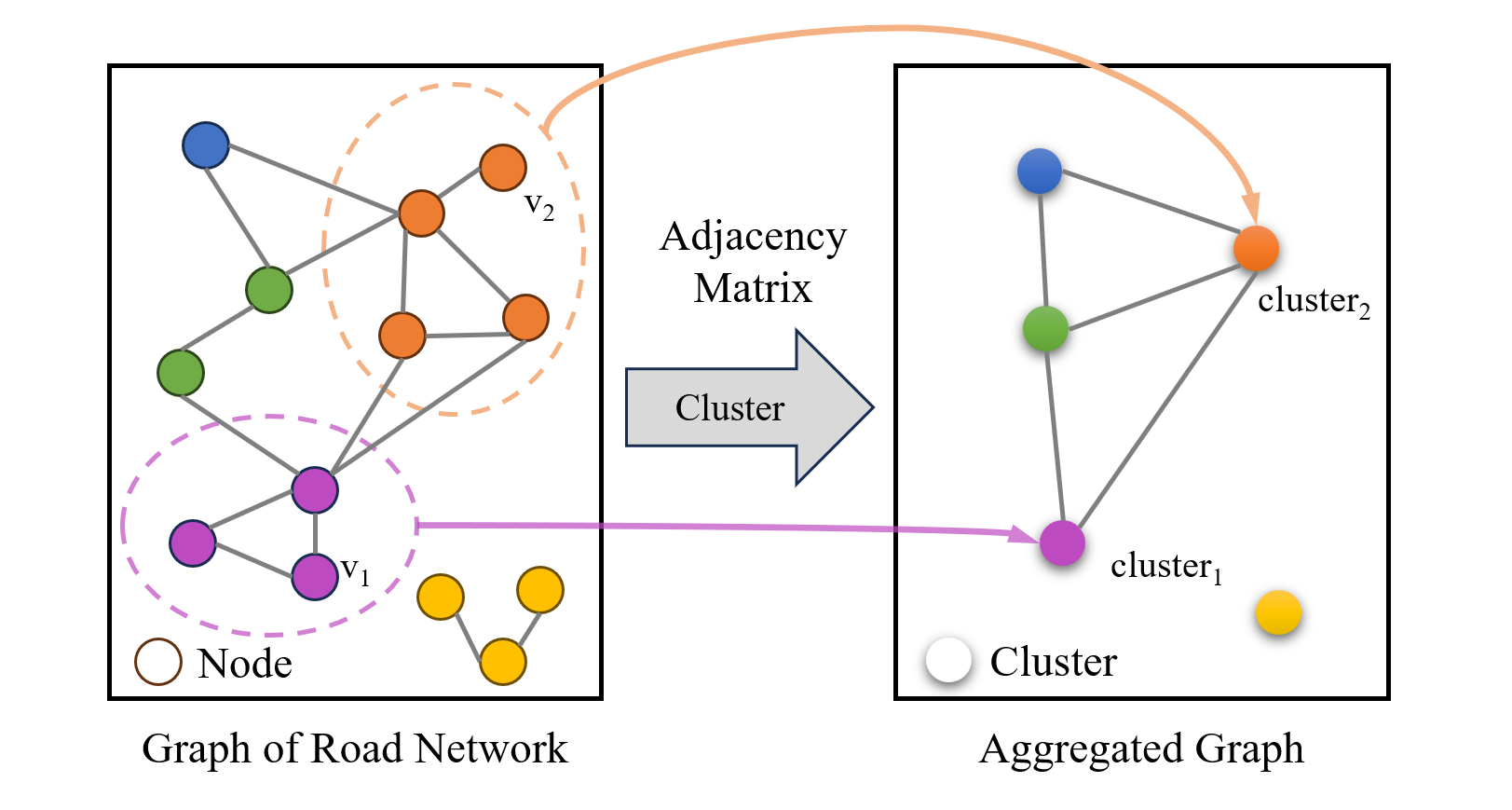}
\caption{The sketch map of the clustering process.}
\Description{A road-network clustering sketch showing how original sensor nodes are partitioned into macro regions.}
\label{fig:clustering_process}
\end{figure}

To obtain the macro graph, we apply spectral clustering to the adjacency matrix of the original road network. As shown in Figure \ref{fig:clustering_process}, the Laplacian of the adjacency matrix is used to partition the road network, and each partition is treated as a macro node. If there exists a node \(v_i \in cluster_1\) and a node \(v_j \in cluster_2\) such that \(v_i\) and \(v_j\) are connected in the original graph, then the corresponding macro nodes are connected in the cluster graph.

Because the road topology is static for a given dataset, the spectral clustering step is executed only once before training and reused throughout all training epochs and test-time predictions. The cluster number \(N^{C}\) controls a trade-off between regional abstraction and local fidelity: a very small \(N^{C}\) tends to over-smooth heterogeneous road segments, whereas an excessively large \(N^{C}\) makes the macro graph too close to the original graph and weakens the efficiency gain. We therefore select \(N^{C}\) according to the validation-set performance for each dataset, and the resulting settings are summarized in Table~\ref{table:cluster_settings}.

\par
The feature of each macro node is obtained by aggregating the average and minimum values of nodes in the same cluster. As a result, the clustered input is represented by \(\bm{X}^{C}=(\bm{X}^{C}_{1},\bm{X}^{C}_{2},...,\bm{X}^{C}_{T_{1}}) \in \mathbb{R}^{N^{C} \times T_{1} \times D}\), where \(N^{C}\) is the number of macro nodes. Thus, the spectral clustering module transforms the original node-level input \(\bm{X}\) into a macro-graph representation \(\bm{X}^{C}\) that is used together with the original branch in the following modules.

\subsection{\bfseries Transmit Blocks}
The transmit block sends region-level trends back to road-segment-level representations, so that each node can simultaneously observe its local state and the macro context of the region to which it belongs. Given \(\bm{X} \in \mathbb{R}^{N \times T_1 \times D}\) and \(\bm{X}^{C} \in \mathbb{R}^{N^{C} \times T_1 \times D}\), the block outputs fused node features \(\bm{\mathcal{X}} \in \mathbb{R}^{N \times T_1 \times 2D}\).

We first define a static assignment matrix \(\bm{Mat}_{T} \in \mathbb{R}^{N \times N^{C}}\) to encode the cluster membership of each node:

\begin{equation}
[\bm{Mat}_{T}]_{ij} =
\begin{cases}
  1 & \text{if node i belongs to cluster j}, \\
  0 & \text{else}.
\end{cases}
\end{equation}

This matrix provides fixed mapping from node space to cluster space. In other words, each row selects the macro node that contains the sensor \(i\).

To adapt this fixed assignment to the time-varying traffic states, we further learn a dynamic transmit matrix \(\bm{Mat}_{T}^{d} \in \mathbb{R}^{N \times N^{C}}\):

\begin{equation}
\begin{aligned}
\bm{F}&=\tanh((\bm{X}^{T}\bm{U}_{1})\bm{U}_{2}(\bm{X}^{C}\bm{U}_{3})^{T}+b) \\
\bm{F}'&=\bm{F}-\text{mean}(\bm{F}) \\
\bm{Mat}_{T}^{d}&=\bm{F}'\bm{Mat}_{T}
\end{aligned}
\end{equation}

The matrix \(\bm{Mat}_{T}^{d}\) reweights the node-to-cluster relations according to the current cross-scale affinity, so that the macro information transmitted can vary with traffic dynamics.

Using \(\bm{Mat}_{T}^{d}\), we project the cluster features back into the node space and concatenate them with the original node features:

\begin{equation}
\begin{aligned}
\bm{\mathcal{X}}^{C} &= \bm{Mat}_{T}^{d} \bm{X}^{C} \\
\bm{\mathcal{X}} &= \bm{X} \oplus \bm{\mathcal{X}}^{C}
\end{aligned}
\end{equation}
Here, \(\bm{\mathcal{X}}^{C} \in \mathbb{R}^{N \times T_1 \times D}\) is the macro feature projected into the node space, and \(\oplus\) denotes the concatenation of the features rather than graph fusion. Therefore, \(\bm{\mathcal{X}}\) retains the local node information and the region-level context simultaneously.

\subsection{\bfseries STF-Blocks(Spatial-Temporal Fusion)}

As shown in Figure \ref{fig:A-STFGCN}(b), each STF-Block contains three tightly coupled components: delay-aware graph construction, gated spatial-temporal fusion graph convolution (GSTF-GCN), and masked temporal self-attention. Given the fused node representation \(\bm{\mathcal{X}}\), this block jointly models local spatial propagation, delay-aware interactions, and long-range temporal dependencies.
\subsubsection{\bfseries Gated Spatial-Temporal Fusion Graph Convolution}\

\begin{figure}[t]
\centering
\includegraphics[width=0.6\linewidth]{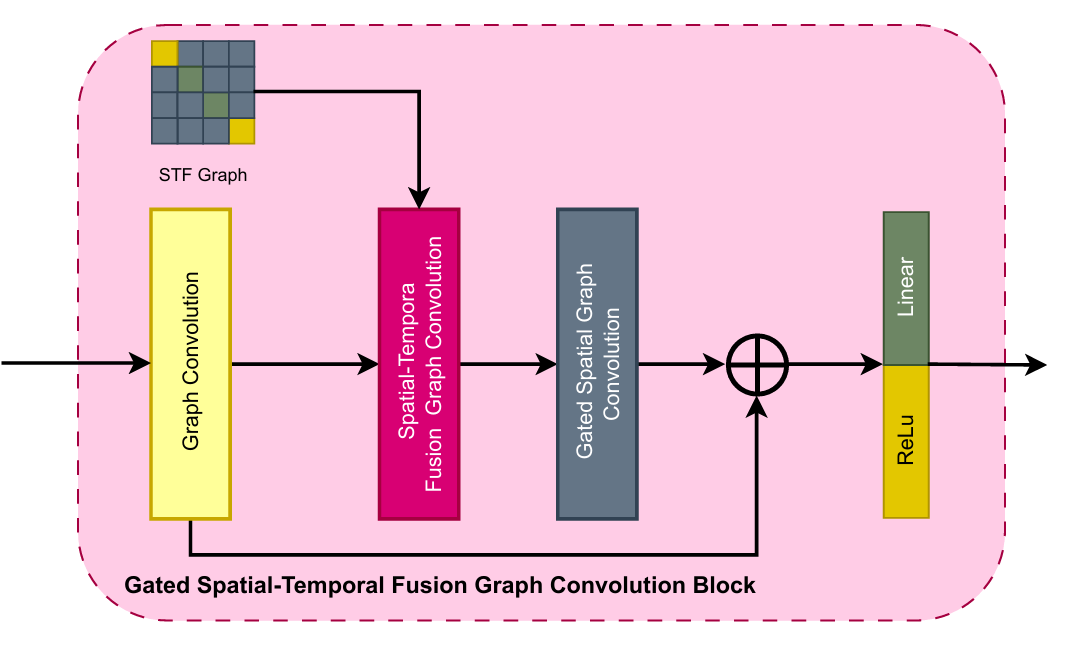}
\caption{Gated Spatial-Temporal Fusion Graph Convolution Block.}
\Description{A block diagram of GSTF-GCN showing graph convolution, spatial-temporal fusion graph convolution, gated temporal convolution, and residual output flow.}
\label{fig:gstf_gcn}
\end{figure}

The internal structure and information flow of GSTF-GCN are illustrated in Figure~\ref{fig:gstf_gcn}.We first construct a spatial-temporal fusion graph to relax the synchronous-propagation assumption made by conventional spatial graphs. In this graph, \(\bm{A}_{sg} \in \mathbb{R}^{N \times N}\) denotes the physical spatial adjacency matrix, \(\bm{A}_{tg}^{1}\) and \(\bm{A}_{tg}^{2}\) denote the strongest 1-step and 2-step delay-temporal links discovered by DTW, and \(\bm{A}_{tc}\) denotes the temporal self-correlation matrix. Their combination forms the spatial-temporal fusion graph \(\bm{A}_{stfg}\), which preserves spatial neighbors, delay-aware cross-node relations, and temporal continuity in a unified structure.

To estimate \(\bm{A}_{tg}^{1}\) and \(\bm{A}_{tg}^{2}\), we use FastDTW to compare delayed node sequences. Figure \ref{fig:dtw_delay_example} gives an intuitive one-step delay example: the downstream series is better aligned after a temporal shift than at the same timestamp, which is exactly the pattern to capture the delay-temporal edges. We restrict the delay-temporal graph to 1-step and 2-step shifts because all datasets are sampled every 5 minutes, so these two graphs correspond to 5-minute and 10-minute propagation lags. For adjacent roads, the dominant spillback or downstream response is usually concentrated in this short window; larger shifts are more easily contaminated by unrelated temporal variations, while their longer-range effects can still be accumulated by stacked STF-Blocks and temporal attention. To keep the graph sparse and stable, we retain only the top-\(k\) delayed neighbors with the smallest alignment distance for each node. The following procedure summarizes this construction:

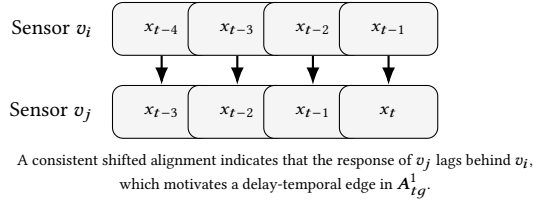
\begin{figure}[t]
\centering
\begin{tikzpicture}[
    >=Latex,
    cell/.style={draw, rounded corners, minimum width=13mm, minimum height=7mm, align=center, font=\scriptsize, fill=gray!8}
]
\node[font=\small, anchor=east] at (0.2,1.8) {Sensor \(v_i\)};
\node[font=\small, anchor=east] at (0.2,0.7) {Sensor \(v_j\)};

\foreach \x/\lab in {1/$x_{t-4}$,2/$x_{t-3}$,3/$x_{t-2}$,4/$x_{t-1}$}
    \node[cell] (a\x) at (\x,1.8) {\lab};
\foreach \x/\lab in {1/$x_{t-3}$,2/$x_{t-2}$,3/$x_{t-1}$,4/$x_{t}$}
    \node[cell] (b\x) at (\x,0.7) {\lab};

\foreach \x in {1,2,3,4}
    \draw[->, thick] (a\x.south) -- (b\x.north);

\node[align=center, font=\scriptsize] at (2.5,-0.15) {A consistent shifted alignment indicates that the response of \(v_j\) lags behind \(v_i\),\\which motivates a delay-temporal edge in \(\bm{A}_{tg}^{1}\).};
\end{tikzpicture}
\caption{Illustrative DTW-based delay alignment between two sensor series. A better alignment after a one-step shift indicates that downstream observations lag behind upstream changes.}
\Description{A two-row timeline diagram for sensors vi and vj. The lower row is shifted one step later, and arrows connect matching pattern positions to illustrate one-step propagation delay.}
\label{fig:dtw_delay_example}
\end{figure}

\setlength{\textfloatsep}{8pt plus 1.0pt minus 1.0pt}
\begin{algorithm}[H]
\caption{Delay-Temporal Graphs Generation}
\label{alg:dtg}
\begin{algorithmic}[1]
\State \textbf{Input:} $\bm{X} = \{ \bm{X}_{1}, \bm{X}_{2}, \dots, \bm{X}_{T_1} \} \in \mathbb{R}^{N \times T_1 \times D}$
\For{$i \gets 1$ to $N$}
    \For{$j \gets 1$ to $N$}
        \State $dist_{i,j}^{1} = \text{TDL}(\bm{X}_{i}[:-1], \bm{X}_{j}[1:])$
        \State $dist_{i,j}^{2} = \text{TDL}(\bm{X}_{i}[:-2], \bm{X}_{j}[2:])$
    \EndFor
    \State Sort smallest $k$ elements and their indices
    \State $\bm{J}^1 = \{ j_1^1, j_2^1, \dots, j_k^1 \}$, $\bm{J}^2 = \{ j_1^2, j_2^2, \dots, j_k^2 \}$
    \For{$j^1$ in $\bm{J}^1$}
        \State $\bm{A_{{tg}_{i,j}}^1} \gets 1$  
        \State $\bm{A_{{tg}_{j,i}}^1} \gets 1$   
    \EndFor
    \For{$j^2$ in $\bm{J}^2$}
        \State $\bm{A_{{tg}_{i,j}}^2} \gets 1$   
        \State $\bm{A_{{tg}_{j,i}}^2} \gets 1$
    \EndFor
\EndFor
\State \textbf{Return:} Weighted Matrices $\bm{A_{tg}}^{1}, \bm{A_{tg}}^{2}$
\end{algorithmic}
\end{algorithm}

For a fixed FastDTW search radius, each pairwise delayed alignment is approximately linear in the sequence length, so constructing \(\bm{A}_{tg}^{1}\) and \(\bm{A}_{tg}^{2}\) is roughly proportional to \(N^{2}T_1\), which is substantially lighter than repeatedly applying standard DTW inside model training. More importantly, this graph construction is performed offline once on the training split, and the resulting sparse graphs are reused during both training and inference. Therefore, the delay-aware modeling improves the receptive graph without introducing per-iteration overhead at deployment time. If the long-term traffic statistics drift, the same offline procedure can be rerun periodically to refresh the delay graph.

Given \(\bm{\mathcal{X}} \in \mathbb{R}^{N \times T_1 \times 2D}\), GSTF-GCN first extracts local topological responses with the graph Laplacian, and then refines them with \(\bm{A}_{stfg}\) to reduce propagation-delay errors:

\begin{equation}
\begin{aligned}
\bm{H}^{1} &= \widetilde{\bm{A}}\bm{\mathcal{X}}\bm{W}_{1}+b_{1} \\
\bm{H}^{2} &= \text{ReLU}(\bm{A}_{stfg}\bm{H}^{1}\bm{W}_{2}+b_{2})
\end{aligned}
\end{equation}
where \(\bm{W}_{1},\bm{W}_{2}\) and \(b_{1},b_{2}\) are trainable parameters, and \(\widetilde{\bm{A}}\) is the Laplacian-based representation corresponding to the graph. Intuitively, \(\bm{H}^{1}\) captures local spatial interactions, while \(\bm{H}^{2}\) further aligns these interactions with delay-aware propagation patterns encoded in \(\bm{A}_{stfg}\).

To extract short-range temporal dynamics efficiently, we further apply a gated dilated convolution to \(\bm{H}^{2}\). This step enlarges the temporal receptive field while keeping the computation lightweight:

\begin{equation}
\begin{aligned}
\left[\bm{\beta}_{1},\bm{\beta}_{2}\right] &= split(Conv_{dil=2}(\bm{H}^{2}))\\
\bm{H}^{3} &= \tanh(\bm{\beta}_{1}) \odot \sigma(\bm{\beta}_{2})
\end{aligned}
\end{equation}
where \(Conv_{dil=2}\) denotes the temporal dilated convolution and split divides the output into two equal parts. The gate selectively preserves useful local temporal responses, and \(\bm{H}^{3} \in \mathbb{R}^{N \times T_1^1 \times D}\), where \(T_1^1=T_1-2*k_c+2\) and \(k_c\) is the temporal kernel size.

To complement the above local modeling, we then apply attention to summarize more global temporal interactions:

\begin{equation}
\begin{aligned}
\bm{H}^{4}&=\bm{W}_3 * \text{Attention}(\bm{X}) \\
&= \frac{\bm{W}_3}{\bm{C}} \sum_{i=1}^{\bm{C}} \left( \bm{H}^{3}_i \cdot \bm{W}_q \cdot (\bm{H}^{3}_i \cdot \bm{W}_k)^T \cdot \bm{H}^{3}_i \cdot \bm{W}_v \right) \\
\bm{H}^{4}&=\bm{H}^{4}+\bm{H}^{3}+b_{3} \\
\bm{GCN}_{output}&=\text{Linear}(\bm{H}^{4} \oplus \bm{H}^{1})
\end{aligned}
\end{equation}
The residual connection preserves local temporal information from \(\bm{H}^{3}\), while the final linear layer fuses global temporal responses with the earlier spatial feature \(\bm{H}^{1}\).

\subsubsection{\bfseries Masked Multi-head Self Attention}\

The second temporal branch of STF-Block is a masked temporal self-attention module, which is used to capture non-local temporal dependencies beyond the convolutional receptive field. Different from a standard causal mask, \(\bm{M}_t\) is a DTW-based temporal similarity mask. Its role is to selectively strengthen interactions between time steps with similar traffic evolution patterns, so that unreliable temporal interactions are suppressed while more trustworthy long-range dependencies are preserved.

For each attention head \(i\), the fused feature \(\bm{\mathcal{X}}\) is projected into query, key, and value tensors:
\begin{equation}
\begin{aligned}
\bm{Q}_i = \bm{\mathcal{X}}\bm{W}_i^Q, \bm{K}_i = \bm{\mathcal{X}}\bm{W}_i^K, \bm{V}_i = \bm{\mathcal{X}}\bm{W}_i^V
\end{aligned}
\end{equation}
where \(\bm{W}_i^Q\), \(\bm{W}_i^K\), and \(\bm{W}_i^V\) are trainable projection matrices. The resulting \(\bm{Q}_i\), \(\bm{K}_i\), and \(\bm{V}_i\) preserve the temporal length \(T_1\) and describe how each time step queries, matches, and aggregates information from the other time steps.

We then perform multi-head self-attention along the temporal dimension:

\begin{equation}
\begin{aligned}
\text{MultiHead}(\bm{\mathcal{X}}) &= \text{Concat}(\text{head}_1, \ldots, \text{head}_h) \bm{W}^O \\
\text{head}_i &= \text{Attention}(\bm{Q}_i, \bm{K}_i, \bm{V}_i) \\
\text{Attention}(\bm{Q}_i, \bm{K}_i, \bm{V}_i) &= \text{softmax}\left(\frac{\bm{Q}_i\bm{K}_i^T}{\sqrt{d_k}}\right) \bm{V}_i
\end{aligned}
\end{equation}

To balance selectivity and completeness, we keep both a masked branch and an unmasked branch. The masked branch emphasizes DTW-consistent temporal alignments, while the unmasked branch preserves global context that may still be useful for prediction:

\begin{equation}
\begin{aligned}
&\bm{MMS}_{output}=\text{Linear}( \\
&(MultiHead(\bm{\mathcal{X}}) \odot \bm{M}_t) \oplus MultiHead(\bm{\mathcal{X}}))
\end{aligned}
\end{equation}
The symbol \(\odot\) denotes the Hadamard product. Therefore, \(\bm{MMS}_{output}\) fuses a DTW-guided temporal representation with an unmasked global temporal representation.

\subsection{\bfseries Output Block}

\noindent To preserve multi-level information from different STF-Blocks, we add a skip connection with a \(1 \times 1\) convolution after each block. This design keeps features from different depths and alleviates information attenuation in deeper layers.

\noindent Specifically, each STF-Block output \(\bm{STF}_{output}^{i}\) is projected to the skip dimension \(\bm{STF}_{sk}^{i} \in \mathbb{R}^{N \times T' \times D_{sk}}\). The skip features are then summed and convolved to obtain the final hidden state \(\bm{\mathcal{X}}_{hid}\), which is mapped to the prediction \(\hat{\bm{Y}} \in \mathbb{R}^{N \times T_2 \times D}\):

\begin{equation}
\begin{aligned}
\bm{STF}_{sk}^i&=Conv(\bm{STF}_{output}^i) \\
\bm{\mathcal{X}}_{hid}&=Conv(\bm{STF}_{sk}^1+\bm{STF}_{sk}^2) \\
\hat{\bm{Y}}&=\text{ReLU}(\text{Linear}(\bm{\mathcal{X}}_{hid}))
\end{aligned}
\end{equation}

\par
We optimize the model using the mean absolute error (MAE). Given the ground truth \(\bm{Y}=(\bm{X}_{T_1+1},\bm{X}_{T_1+2},...,\bm{X}_{T_1+T_2})\), the loss is defined as

\begin{equation}
MAE(\bm{Y},\hat{\bm{Y}}) = \frac{1}{N} \sum_{i=1}^{N} |\hat{\bm{Y}}_i - \bm{Y}_i|
\end{equation}

\begin{table*}[!h]
\centering
\captionsetup{justification=centering, width=\textwidth}
\caption{Experimental Results on PeMS04, PeMS07, and PeMS08.}
\label{table:PeMS}
{\tiny
\begin{tabularx}{\textwidth}{Z|Z|ZZZ|ZZZ|ZZZ}
\toprule
\multicolumn{2}{c|}{\textbf{Datasets}} & \multicolumn{3}{c|}{\textbf{PEMS04}} & \multicolumn{3}{c|}{\textbf{PEMS07}} & \multicolumn{3}{c}{\textbf{PEMS08}} \\
\midrule
\multicolumn{2}{c|}{\textbf{Metrics}} & \textbf{MAE} & \textbf{RMSE} & \textbf{Accuracy} & \textbf{MAE} & \textbf{RMSE} & \textbf{Accuracy} & \textbf{MAE} & \textbf{RMSE} & \textbf{Accuracy} \\
\midrule
\multirow{9}{*}{\textbf{Models}}&ARIMA & 30.792 & 36.106 & 0.757 & 36.761 & 44.952 & 0.719 & 25.817 & 36.204 & 0.868 \\
&GRU & 28.469 & 41.322 & 0.836 & 29.236 & 43.504 & 0.830 & 22.852 & 35.023 & 0.877 \\ 
&STGCN & 27.190 & 34.166 & 0.851 & 30.108 & 44.050 & 0.833 & 20.125 & 32.589 & 0.878 \\
&T-GCN & 28.963 & 44.487 & 0.835 & 28.624 & 41.368 & 0.832 & 20.717 & 33.004 & 0.883 \\ 
\cmidrule(lr){2-11}
&STFGNN & 21.045 & 33.003 & \textbf{0.878} & 24.015 & 37.011 & 0.869 & 17.636 & 25.985 & 0.906 \\
&ASTGCN & 22.175 & 34.166 & 0.873 & 26.748 & 40.817 & 0.868 & 16.824 & 25.892 & 0.908 \\
&STAGCN{-}EC & 22.445 & 33.689 & 0.876 & 26.236 & 37.811 & 0.869 & --- & --- & --- \\ 
&HGCN & 22.210 & 33.823 & 0.872 & 29.052 & 39.942 & 0.872 & 17.428 & 26.122 & 0.908 \\ 
\cmidrule(lr){2-11}
&A-STFGCN & \textbf{20.515} & \textbf{31.974} & \textbf{0.878} & \textbf{22.208} & \textbf{34.599} & \textbf{0.874} & \textbf{16.018} & \textbf{25.465} & \textbf{0.924} \\
\bottomrule 
\end{tabularx}
}
\end{table*}

\begin{table*}[!h]
\centering
\captionsetup{justification=centering, width=\textwidth}
\caption{Experimental Results on CA and GLA.}
\label{table:CA & GLA}
\begin{tabularx}{\textwidth}{Z|Z|ZZZ|ZZZ}
\toprule
\multicolumn{2}{c|}{\textbf{Datasets}} & \multicolumn{3}{c|}{\textbf{CA}} & \multicolumn{3}{c}{\textbf{GLA}} \\
\midrule
\multicolumn{2}{c|}{\textbf{Metrics}} & \textbf{MAE} & \textbf{RMSE} & \textbf{Accuracy} & \textbf{MAE} & \textbf{RMSE} & \textbf{Accuracy} \\
\midrule
\multirow{8}{*}{\textbf{Models}} & ARIMA & 37.052 & 52.114 & 0.754 & 40.704 & 57.861 & 0.702 \\
& GRU & 30.178 & 44.747 & 0.798 & 32.246 & 44.650 & 0.772 \\ 
& STGCN & 21.131 & 36.792 & 0.862 & 22.687 & 38.102 & 0.857 \\
& T-GCN & 26.967 & 39.667 & 0.842 & 27.052 & 42.712 & 0.841 \\ 
\cmidrule(lr){2-8}
& STFGNN & 21.032 & 35.580 & 0.868 & 21.648 & 36.785 & 0.872 \\
& ASTGCN & 28.052 & 44.623 & 0.820 & 27.989 & 42.783 & 0.822 \\
& GWNET & 21.515  & 34.024 & 0.871 & \textbf{20.232} & 32.886 & \textbf{0.883} \\ 
& HGCN & 20.986 & 33.475 & 0.881 & 21.015 & 33.768 & 0.878 \\ 
\cmidrule(lr){2-8}
& A-STFGCN & \textbf{19.883} & \textbf{31.213} & \textbf{0.886} & 20.252 & \textbf{32.526} & \textbf{0.883} \\
\bottomrule 
\end{tabularx}
\end{table*}

\section{Experiments}
In this section, we provide five traffic flow datasets to evaluate the method, then compare the proposed A-STFGCN with state-of-the-art existing work, and finally analyze the effectiveness of A-STFGCN through ablation studies.

\subsection{\bfseries Datasets}

We conducted our model evaluation using five highway traffic datasets: PeMS07, PeMS04, PeMS08, CA, and GLA. PeMS04, PeMS07, and PeMS08 are derived from real-time data collected by the California Department of Transportation's Performance Measurement System (PeMS). After preprocessing, they provide traffic data every five minutes. In the experiment, we used traffic flow data from the past hour to predict traffic flow for the next hour. These three datasets contain distance information from adjacent sensors, so we can extract spatial features from both adjacent and distance information.
 CA and GLA are part of the LargeST benchmark dataset. CA includes a total of 8,600 sensors with data spanning five years (from 2017 to 2021), covering a wide area of California. GLA, a subset of CA, includes 3,834 sensors installed in 5 counties in the Greater Los Angeles area. The LargeST dataset provides comprehensive metadata for each sensor, including coordinates, highway category, number of lanes, and more, allowing detailed spatial-temporal analysis. Similar to PeMS datasets, CA and GLA provide traffic data every five minutes.
 Table \ref{table:datasets} meticulously details the dataset specifications. We employ Z-score normalization for data processing from all datasets. We construct the adjacency matrix of the road network. Each dataset is divided into 60\% for training, 20\% for validation, and 20\% for testing in chronological order.

\begin{table}[H]
\centering
\caption{Description of traffic datasets used in this study.}
\label{table:datasets}
\begin{tabular}{@{\extracolsep{\fill}}c|c|c|c|c}
\toprule
\textbf{Datasets} & \textbf{\#Nodes} & \textbf{\#Edges} & \textbf{\#Timesteps} & \textbf{\#Interval(min)}\\
\midrule
PeMS04 & 307 & 340 & 16,992 & 5\\
PeMS07 & 883 & 866 & 28,224 & 5\\ 
PeMS08 & 170 & 290 & 17,856 & 5\\ 
CA & 8600 & 201363 & 525,888 & 5\\
GLA & 3834 & 98703 & 525,888 & 5\\ 
\bottomrule 
\end{tabular}
\end{table}

\subsection{\bfseries Evaluation}
In our experimental analysis, we assess performance using three pivotal metrics: (1) Mean Absolute Error (MAE), (2) Root Mean Squared Error (RMSE), and (3) Accuracy. It is important to note that missing values within the datasets have been excluded from these calculations to ensure the integrity and accuracy of our metric evaluations.

\subsection{\bfseries Experimental Settings}
All experiments in this work were carried out using PyTorch 1.2.0 on a workstation equipped with an NVIDIA RTX 3090 GPU (24 GB of memory). The best checkpoint was selected according to the validation-set performance, and the model was optimized with Adam using an initial learning rate of 0.0001. We trained the model for 80 epochs with a batch size of 32.
\par
For all five datasets, the sampling interval is 5 minutes. Therefore, using the past hour to predict the next hour corresponds to the input steps \(T_1=12\) and the prediction steps \(T_2=12\) . The delay-temporal graph \(A_{TG}\) in Algorithm~\ref{alg:dtg} was generated offline before model training and then reused during both training and inference. In our implementation, this preprocessing step took around 20 minutes for most public datasets. The FastDTW search length was set to 12, which is consistent with the longest forecasting horizon considered in this work. We use only 1-step and 2-step delay graphs because they correspond to the dominant 5-minute and 10-minute local propagation lags under our sampling rate, while longer effective dependencies are further modeled by stacked STF-Blocks and temporal attention. The sparsity parameter \(k\) in Algorithm~\ref{alg:dtg} and the cluster number \(N^{C}\) were selected based on the validation-set performance. To simplify graph fusion, the entries of \(\bm{A}_{sg}\), \(\bm{A}_{tg}^{1}\), \(\bm{A}_{tg}^{2}\), and \(\bm{A}_{tc}\) were binarized to \(\{0,1\}\).

\begin{table*}[!t]
\centering
\caption{Reproducibility summary of A-STFGCN.}
\label{table:hyperparameters}
\small
\begin{tabularx}{\textwidth}{l l X}
\toprule
\textbf{Category} & \textbf{Parameter} & \textbf{Value} \\
\midrule
Data setting & Sampling interval & 5 minutes for all datasets. \\
Data setting & Input / prediction horizon & \(T_1=12\) and \(T_2=12\), corresponding to past 60 minutes for future 60-minute prediction. \\
Data setting & Data split & Chronological split with 60\% training, 20\% validation, and 20\% testing. \\
Data setting & Normalization & Z-score normalization. \\
Training & Framework / hardware & PyTorch 1.2.0 on an NVIDIA RTX 3090 GPU with 24 GB memory. \\
Training & Optimizer / learning rate & Adam with initial learning rate \(1\times 10^{-4}\). \\
Training & Batch size / epochs & Batch size \(=32\); training epochs \(=80\). \\
Delay graph & Delay steps & 1-step and 2-step delay-temporal graphs, i.e., \(\bm{A}_{tg}^{1}\) and \(\bm{A}_{tg}^{2}\). \\
Delay graph & FastDTW search length & 12. \\
Delay graph & Construction mode & Precomputed offline before training; typical preprocessing time is about 20 minutes on public datasets. \\
Architecture & STF-Block temporal dilation & \(dil=2\). \\
Architecture & Temporal kernel size & \(t_s=1\). \\
Architecture & Graph value type & The entries of the fused subgraphs are binarized to \(\{0,1\}\). \\
Architecture & Output skip mapping & \(1 \times 1\) convolution after each STF-Block. \\
\bottomrule
\end{tabularx}
\end{table*}

\begin{table}[!t]
\centering
\caption{Dataset-specific cluster settings.}
\label{table:cluster_settings}
\small
\begin{tabular}{l c}
\toprule
\textbf{Dataset} & \(\bm{N^C}\) \\
\midrule
PeMS04 & 20 \\
PeMS07 & 40 \\
PeMS08 & 40 \\
CA & 400 \\
GLA & 200 \\
\bottomrule
\end{tabular}
\end{table}

The dataset-specific cluster counts are summarized in Table~\ref{table:cluster_settings}. In particular, we set \(N^C=400\) for CA, \(N^C=200\) for GLA, \(N^C=20\) for PeMS04, and \(N^C=40\) for PeMS07 based on validation-set performance. For datasets not explicitly listed in the tuning record, we followed the same validation-based selection protocol.

\subsection{\bfseries Baselines}
The proposed model A-STFGCN is compared with eight well-known baseline models as follows:

\par
\textbf{ARIMA}: The Auto-Regressive Integrated Moving Average model is a classical statistical approach used for predicting future values in time-series data, leveraging dependencies within the series.
\par
\textbf{GRU}: The Gated Recurrent Unit network, which represents an enhancement over the traditional Recurrent Neural Network (RNN), aims to solve the vanishing gradient problem commonly associated with standard RNNs\cite{cho2014properties}.
\par
\textbf{STGCN}: Spatial-Temporal Graph Convolutional Networks incorporate both Gated Linear Unit (GLU) and graph convolution operations to address spatial and temporal data dependencies\cite{yu2017spatio}.  
\par
\textbf{T-GCN}: The Temporal Graph Convolutional Network is designed to integrate temporal dynamics with graph convolutional methods to enhance prediction accuracy in network data\cite{zhao2019t}.
\par
\textbf{STFGNN}: The Spatial-Temporal Fusion Graph Neural Network utilizes the DTW algorithm to generate the temporal graph and designs the spatial-temporal fusion graph neural module to synchronously capture the spatial-temporal relation\cite{li2021spatial}.
\par
\textbf{ASTGCN}: Attention Based Spatial-Temporal Graph Convolutional Network combines attention mechanisms with the characteristics of spatial-temporal data\cite{guo2019attention}.
\par
\textbf{HGCN}: A novel Hierarchical Graph Convolution Networks for traffic forecasting by operating on both the micro and macro traffic graphs\cite{guo2021hierarchical}.
\par
\textbf{STAGCN-EC}: A Spatial-Temporal Attention Graph Convolution Network on Edge Cloud model, model training requires low computational resources and is efficient\cite{lai2022spatial}.

\subsection{\bfseries Experiment Results and Analysis}

Table \ref{table:PeMS} and Table \ref{table:CA & GLA} illustrate the average accuracy comparison of traffic flow predictions at 12 time steps (1 hour) between the proposed A-STFGCN model and the eight baseline models mentioned above on five datasets used. Models that perform poorly on the respective datasets are not included in the comparison.

\par
Traditional speed prediction methods mainly model temporal regularity and do not explicitly exploit the irregular spatial correlations of road networks. ARIMA therefore performs poorly, while GRU improves temporal modeling but still lacks graph-based spatial propagation. The graph-based baselines, including STGCN, T-GCN, STFGNN, ASTGCN, HGCN, STAGCN-EC, and A-STFGCN, achieve clear improvements because they encode spatial-temporal dependencies. Among them, STFGNN explores potential temporal correlations with FastDTW, ASTGCN and STAGCN-EC introduce attention-based temporal modeling, and HGCN incorporates hierarchical regional information. A-STFGCN further combines these useful directions by using a macro graph, a transmit block, delay-aware fusion graph convolution, and masked temporal self-attention.

\par
A detailed analysis of the experimental results shows that A-STFGCN achieves consistently competitive MAE and RMSE across the five datasets. On PeMS04, A-STFGCN reduces MAE by 7.49\% and RMSE by 5.10\% compared with the strongest baseline, indicating improved numerical prediction reliability even when the accuracy metric is close. On PeMS07, A-STFGCN obtains a larger improvement, with MAE reduced by 15.35\%, RMSE reduced by 8.49\%, and accuracy increased by 1.57\%. PeMS07 contains many more sensors than PeMS04 and PeMS08, so this result suggests that the macro-micro feature interaction and delay-aware graph are particularly useful when the road network becomes larger and spatial propagation is more heterogeneous. On PeMS08, the model still outperforms all baselines, although the margin is smaller because the dataset contains fewer sensors and the spatial structure is less complex.
\par
The same trend can be observed on the LargeST subsets. For CA, A-STFGCN achieves the lowest MAE and RMSE, reducing MAE and RMSE by 5.26\% and 6.75\% compared with the next best model HGCN, while also improving accuracy by 0.57\%. CA contains 8,600 sensors and covers a much larger road network, where relying only on local topology can miss long-range and region-level propagation. The cluster graph and transmit block provide stable regional context, while the delay-temporal graph helps align cross-node responses under short propagation lags. For GLA, A-STFGCN obtains the lowest RMSE of 32.526 and an accuracy of 0.883, which is comparable to the best baseline. Although its MAE is the second lowest, the lower RMSE indicates that A-STFGCN is more robust to large prediction errors and peak deviations, which are important in complex urban traffic conditions.
\par
Following common practice in the compared baselines, we report the results selected by validation-set performance. We did not add a separate multi-run standard deviation table in this revision because rerunning all baselines under the same protocol would require substantial additional experiments. We therefore focus the analysis on the observed performance trends and provide more detailed mechanism-level interpretation of the results.

\begin{table}[H]
\begin{center}
\caption{Comparison of average training time (in hours) on PeMS04 and PeMS07 datasets.}
\label{table:trainingTime}
\begin{tabular}{@{\extracolsep{\fill}}c|ccc}
\toprule
\multirow{2}{*}{\textbf{Datasets}} & \textbf{Training Time(h)} \\
\cmidrule(lr){2-4}
& \textbf{A-STFGCN} & \textbf{STAGCN-EC} & \textbf{HGCN}\\
\midrule
PeMS04 & 1.21 & 1.25 & 0.91 \\
PeMS07 & 3.61 & 3.53 & 5.13 \\ 
\bottomrule 
\end{tabular}
\end{center}
\end{table}
\par
To achieve higher prediction accuracy, some baseline models rely on complex structures that require more time for training and occupy more GPU memory, especially on larger datasets. Excessive training and prediction time can limit practical deployment in urban transportation systems. Table \ref{table:trainingTime} compares A-STFGCN with STAGCN-EC and HGCN, two representative baselines that also consider computational efficiency. A-STFGCN remains close to STAGCN-EC on PeMS04 and PeMS07 while being faster than HGCN on PeMS07, showing that the proposed macro-micro and delay-aware design does not introduce prohibitive training overhead. Moreover, the spectral clustering and delay-temporal graph construction are performed offline and reused during training and inference, so they do not increase the per-epoch or online prediction cost.

\subsection{\bfseries Ablation Study}
\begin{figure}[!ht]
\centering
\includegraphics[width=0.7\linewidth]{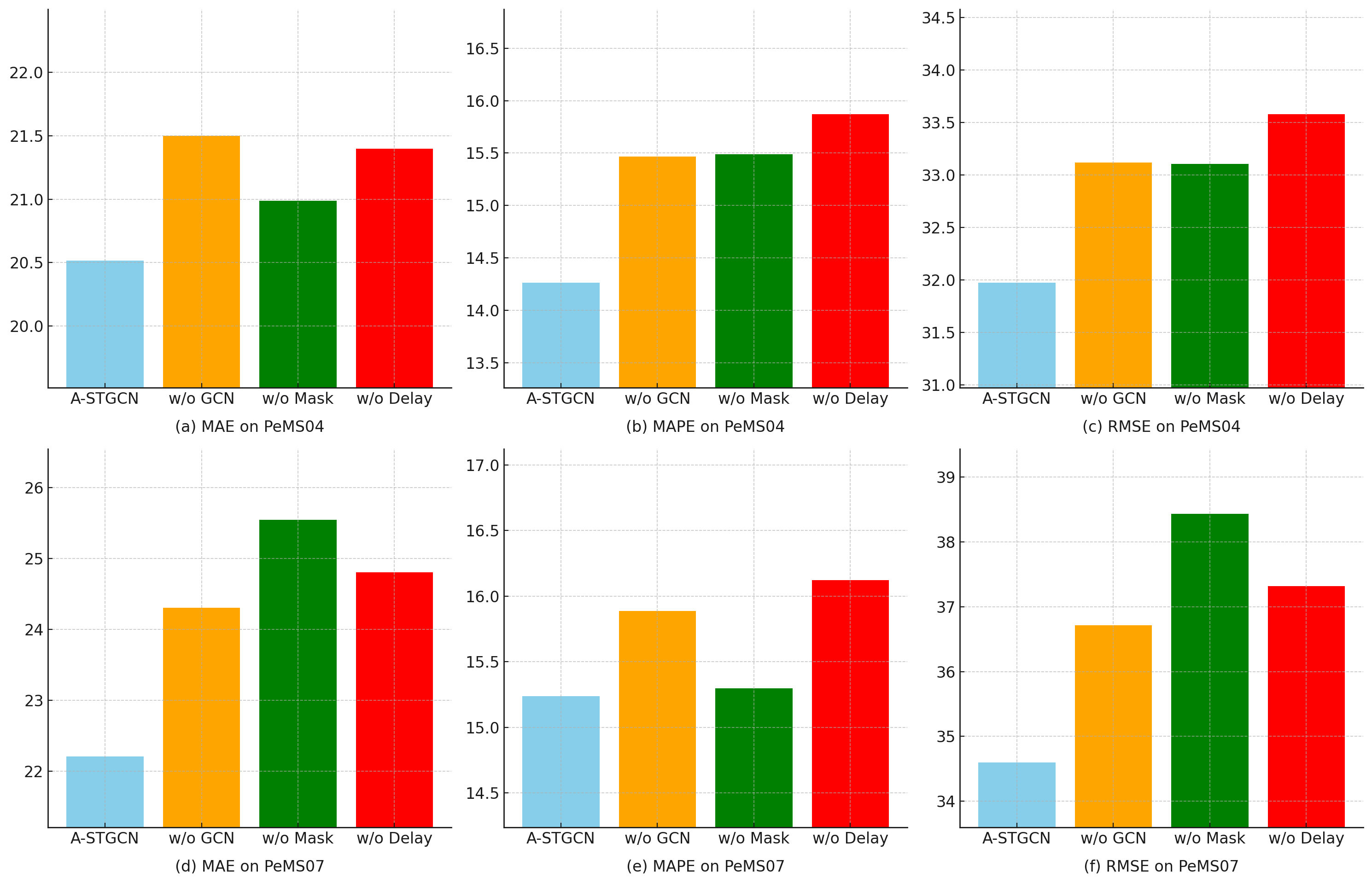}
\caption{Ablation Study on PeMS04 and PeMS07. Each bar represents MAE performance under different ablation settings.}
\Description{A bar chart comparing A-STFGCN with ablated variants on PeMS04 and PeMS07 using MAE.}
\label{fig:Ablation}
\end{figure}

\begin{figure}[!ht]
\centering
\includegraphics[width=0.6\linewidth]{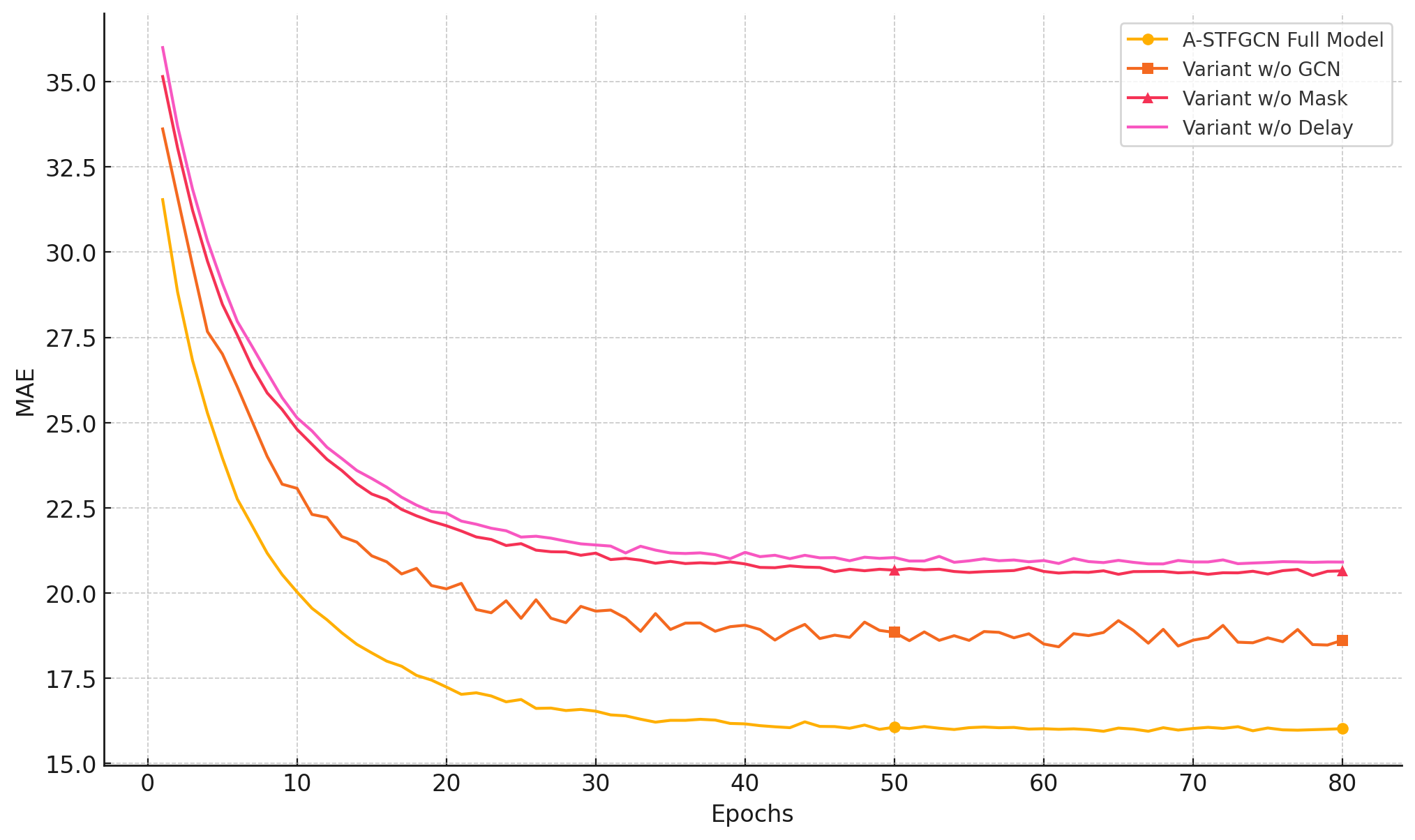}
\caption{MAE trend over training epochs on the PeMS08 dataset. The A-STFGCN shows stable convergence compared to baselines.}
\Description{A line chart showing the MAE trend over training epochs on the PeMS08 dataset.}
\label{fig:Ablation-MAE}
\end{figure}
We conducted comparative experiments on three variants to analyze the effectiveness of different structures in A-STFGCN. (1) w/o GCN replaces the GSTF-GCN module with regular graph convolutional layers. This variant removes message passing on the spatial-temporal fusion graph and therefore weakens the ability to jointly exploit physical adjacency, temporal self-correlation, and delay-temporal links. (2) w/o mask removes the DTW-guided mask from the temporal multi-head self-attention module. Without this selective temporal filtering, the attention branch treats all temporal interactions more uniformly and may emphasize less reliable temporal alignments. (3) w/o Delay replaces the delay-temporal graph with a synchronous time-correlation matrix. This setting keeps temporal correlation modeling but removes explicit propagation-lag compensation, so it directly tests whether shifted DTW alignment contributes to prediction.

\par
Figure \ref{fig:Ablation} compares these variants on the PeMS04 and PeMS07 datasets. The w/o GCN variant performs worse because regular graph convolution cannot fully exploit the fused delay-aware spatial-temporal graph. The w/o Delay variant also degrades, confirming that synchronous correlations alone are insufficient when traffic perturbations propagate across sensors with short time lags. The w/o mask variant shows that temporal self-attention benefits from DTW-guided selectivity rather than attending to all time steps uniformly. Overall, the ablation results indicate that A-STFGCN gains from the cooperation of the fusion graph convolution, the delay-temporal graph, and the masked temporal self-attention module.

\section{Limitations and Future Work}
Although A-STFGCN improves the modeling of spatial-temporal propagation in traffic networks, several limitations remain. First, the spectral cluster graph and the delay-temporal graph are constructed offline from historical topology and traffic sequences. This design keeps online training and inference efficient, but the graphs may need to be periodically updated when the road network, sensor deployment, or long-term traffic pattern changes substantially. Second, the cluster number $N^C$ and the sparsity of delay-temporal links are selected according to the validation performance. A fixed granularity is simple and reproducible, but may be suboptimal for cities with highly heterogeneous regional structures. Future work can explore adaptive multi-resolution clustering and dynamic update strategies while preserving the interpretability of the macro graph. Third, the current model relies mainly on historical traffic observations and graph topology. It does not explicitly incorporate external factors such as weather, holidays, incidents, or planned events, which may cause abrupt non-periodic fluctuations. Integrating such exogenous signals with the delay-aware graph is a promising direction to improve robustness under abnormal traffic conditions. Finally, we plan to extend A-STFGCN to other spatial-temporal prediction tasks, such as rainfall prediction and accident prediction, and investigate pre-training or data augmentation techniques for scenarios with insufficient temporal data.

\section{Conclusion}
This study presents a novel network model, the Attention-Based Spatial-Temporal Fusion Graph Convolution Network (A-STFGCN), to predict traffic flow. Traffic flow prediction is crucial for optimizing transportation systems, reducing congestion, and enhancing overall urban mobility. Although various graph convolution-based models have been proposed to extract spatial-temporal features and predict traffic flow, most of them emphasize static spatial dependencies and spatial-temporal relationships in topological structures, neglecting spatial-temporal heterogeneity issues.

Our proposed A-STFGCN model addresses these challenges by introducing a Spatial-Temporal Fusion Block to extract spatial-temporal feature correlations. We leverage a Mask matrix-based multi-head self-attention mechanism to capture long-term and short-term temporal features of the data. Extensive experiments on five real-world datasets demonstrate that our method not only achieves the overall best performance, but also exhibits excellent computational efficiency and data utilization efficiency compared to the eight baseline methods.

\section{Acknowledgements}
This work was supported in part by the Beijing Natural Science Foundation(Grant No.L233034), in part by the National Natural Science Foundation of China (No.62572075), in part by Fundamental Research Funds for the Beijing University of Posts and Telecommunications (No.2025TSQY01), and in part by Hubei Key Laboratory of Intelligent Robot (Wuhan Institute of Technology)(No.HBIR202302).

\bibliographystyle{ACM-Reference-Format}
\bibliography{references}

\end{document}